\documentclass[conference]{IEEEtran}
\IEEEoverridecommandlockouts
\usepackage{cite}
\usepackage{amsmath,amssymb,amsfonts}
\usepackage{algorithmic}
\usepackage{graphicx}
\usepackage{textcomp}
\usepackage{xcolor}
\usepackage[ruled,linesnumbered]{algorithm2e}
\usepackage{multirow}
\usepackage{enumerate}
\usepackage{url}
\usepackage{booktabs}
\usepackage{threeparttable}
\usepackage{endnotes}
\usepackage{bbding}
\usepackage{bm}
\usepackage{CJKutf8}
\usepackage[utf8]{inputenc}
\usepackage{subfigure}

\def\BibTeX{{\rm B\kern-.05em{\sc i\kern-.025em b}\kern-.08em
    T\kern-.1667em\lower.7ex\hbox{E}\kern-.125emX}}

\begin{document}
\begin{CJK}{UTF8}{gbsn}
\title{Question Answering based Clinical Text Structuring Using Pre-trained Language Model}

\author{\IEEEauthorblockN{Jiahui Qiu$^1$, Yangming Zhou$^{1,*}$, Zhiyuan Ma$^{1}$, Tong Ruan$^{1,*}$, Jinlin Liu$^1$, and Jing Sun$^2$}
\IEEEauthorblockA{$^1$School of Information Science and Engineering, East China University of Science and Technology, Shanghai 200237, China \\
$^2$Ruijin Hospital, School of Medicine, Shanghai Jiao Tong University, Shanghai 200025, China \\
$^*$Corresponding authors\\
Emails: \{ymzhou,ruantong\}@ecust.edu.cn\\
}
}

\maketitle

\begin{abstract}

Clinical text structuring is a critical and fundamental task for clinical research. Traditional methods such as task-specific end-to-end models and pipeline models usually suffer from the lack of dataset and error propagation. In this paper, we present a question answering based clinical text structuring (QA-CTS) task to unify different specific CTS tasks and make dataset shareable. A novel model that aims to introduce domain-specific features (e.g., clinical named entity information) into pre-trained language model is also proposed for QA-CTS task. Experimental results on Chinese pathology reports collected from Ruijing Hospital demonstrate our presented QA-CTS task is very effective to improve the performance on specific tasks. Our proposed model also competes favorably with strong baseline models in specific tasks.

\end{abstract}

\begin{IEEEkeywords}

Question answering, Clinical text structuring, Pre-trained language model, Electronic health records.

\end{IEEEkeywords}

\section{Introduction}
\label{Sec:Introduction}

Clinical text structuring (CTS) is a critical task for fetching medical research data from electronic health records (EHRs), where structural patient medical data, such as whether the patient has specific symptoms, diseases, or what the tumor size is are obtained. It is important to extract structured data from clinical text because bio-medical systems or bio-medical researches greatly rely on structured data but they cannot obtain them directly. In addition, clinical text often contains abundant healthcare information. CTS is able to provide large-scale extracted structured data for enormous down-stream clinical researches.

However, end-to-end CTS is a very challenging task. Different CTS tasks often have non-uniform output formats, such as specific-class classifications (e.g. tumor stage), strings in the original text (e.g. result for a laboratory test) and inferred values from part of the original text (e.g. calculated tumor size). Researchers have to construct different models for it, which is already costly, and hence it calls for a lot of labeled data for each model. Moreover, labeling necessary amount of data for training neural network requires expensive labor cost. To handle it, researchers turn to some rule-based structuring methods which often have lower labor cost.

Traditionally, CTS tasks can be addressed by rule and dictionary based methods~\cite{fukuda1998toward,wang2006linguistic, song2015developing}, task-specific end-to-end methods~\cite{topaz2016automated,tan2016development,senders2019natural} and pipeline methods~\cite{bill2014automated,iqbal2017adept,fonferko2019using}. Rule and dictionary based methods suffer from costly human-designed extraction rules, while task-specific end-to-end methods have non-uniform output formats and require task-specific training dataset. Pipeline methods break down the entire process into several pieces which improves the performance and generality. However, when the pipeline depth grows, error propagation will have a greater impact on the performance.

\begin{figure}[!ht]
\begin{center}
\includegraphics[width=3.5in]{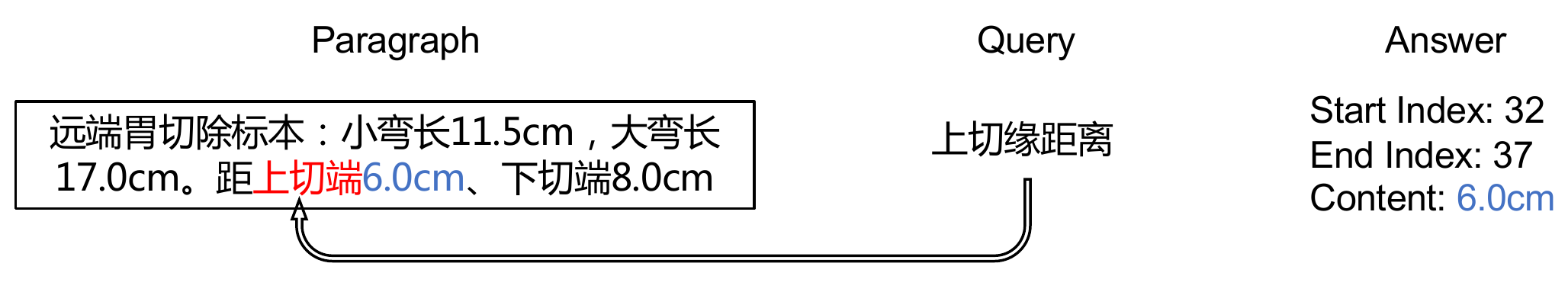}
\caption{An illustrative example of QA-CTS task.}
\label{Fig:Task}
\end{center}
\end{figure}

To reduce the pipeline depth and break the barrier of non-uniform output formats, we present a question answering based clinical text structuring (QA-CTS) task (see Fig.~\ref{Fig:Task}). Unlike the traditional CTS task, our QA-CTS task aims to discover the most related text from original paragraph text. For some cases, it is already the final answer in deed (e.g., extracting sub-string). While for other cases, it needs several steps to obtain the final answer, such as entity names conversion and negative words recognition. Our presented QA-CTS task unifies the output format of the traditional CTS task and make the training data shareable, thus enriching the training data. The main contributions of this work can be summarized as follows.
\begin{itemize}
    \item We first present a question answering based clinical text structuring (QA-CTS) task, which unifies different specific tasks and make dataset shareable. We also propose an effective model to integrate clinical named entity information into pre-trained language model.
    \item Experimental results show that QA-CTS task leads to significant improvement due to shared dataset. Our proposed model also achieves significantly better performance than the strong baseline methods. In addition, we also show that two-stage training mechanism has a great improvement on QA-CTS task.
\end{itemize}

\section{Related Work}
\label{Sec:Related Work}

\subsection{Clinical Text Structuring}
Clinical text structuring is an important problem which is highly related to practical applications. Considerable efforts have been  made on CTS task. These studies can be roughly divided into three categories, namely rule and dictionary based methods, task-specific end-to-end methods and pipeline methods. Rule and dictionary based methods~\cite{fukuda1998toward,wang2006linguistic,song2015developing} rely extremely on heuristics and handcrafted extraction rules which is more of an art than a science and incurring extensive trial-and-error experiments. Task-specific end-to-end methods~\cite{topaz2016automated,tan2016development} use large amount of data to automatically model the specific task. However, none of their models could be used to another task due to output format difference. This makes building a new model for a new task a costly job. Pipeline methods~\cite{bill2014automated,iqbal2017adept,fonferko2019using} break down the entire task into several basic natural language processing tasks. This kind of method focus on language itself, so it can handle tasks more general. However, as the depth of pipeline grows, it is obvious that error propagation will be more and more serious. In contrary, using less components to decrease the pipeline depth will lead to a poor performance. So the upper limit of this method depends mainly on the worst component.

\subsection{Pre-trained Language Model}

Recently, some studies focused on using pre-trained language representation models to capture language information from text and then utilizing the information to improve the performance of specific natural language processing (NLP) tasks~\cite{radford2018improving,peters-etal-2018-deep,devlin2018bert,yang2019xlnet} which makes language model a shared model to all NLP tasks. The main motivation of introducing pre-trained language model is to solve the shortage of labeled data and polysemy problem. Although polysemy problem is not a common phenomenon in biomedical domain, shortage of labeled data is always a non-trivial problem. Lee et al.~\cite{lee2019biobert} applied BERT on large-scale biomedical unannotated data and achieved improvement on biomedical named entity recognition, relation extraction and question answering. Kim et al.~\cite{Kimn2019neural} adapted BioBERT into multi-type named entity recognition and discovered new entities. Both of them demonstrates the usefulness of introducing pre-trained language model into biomedical domain.

\begin{figure}[!ht]
\begin{center}
\includegraphics[width=3.0in]{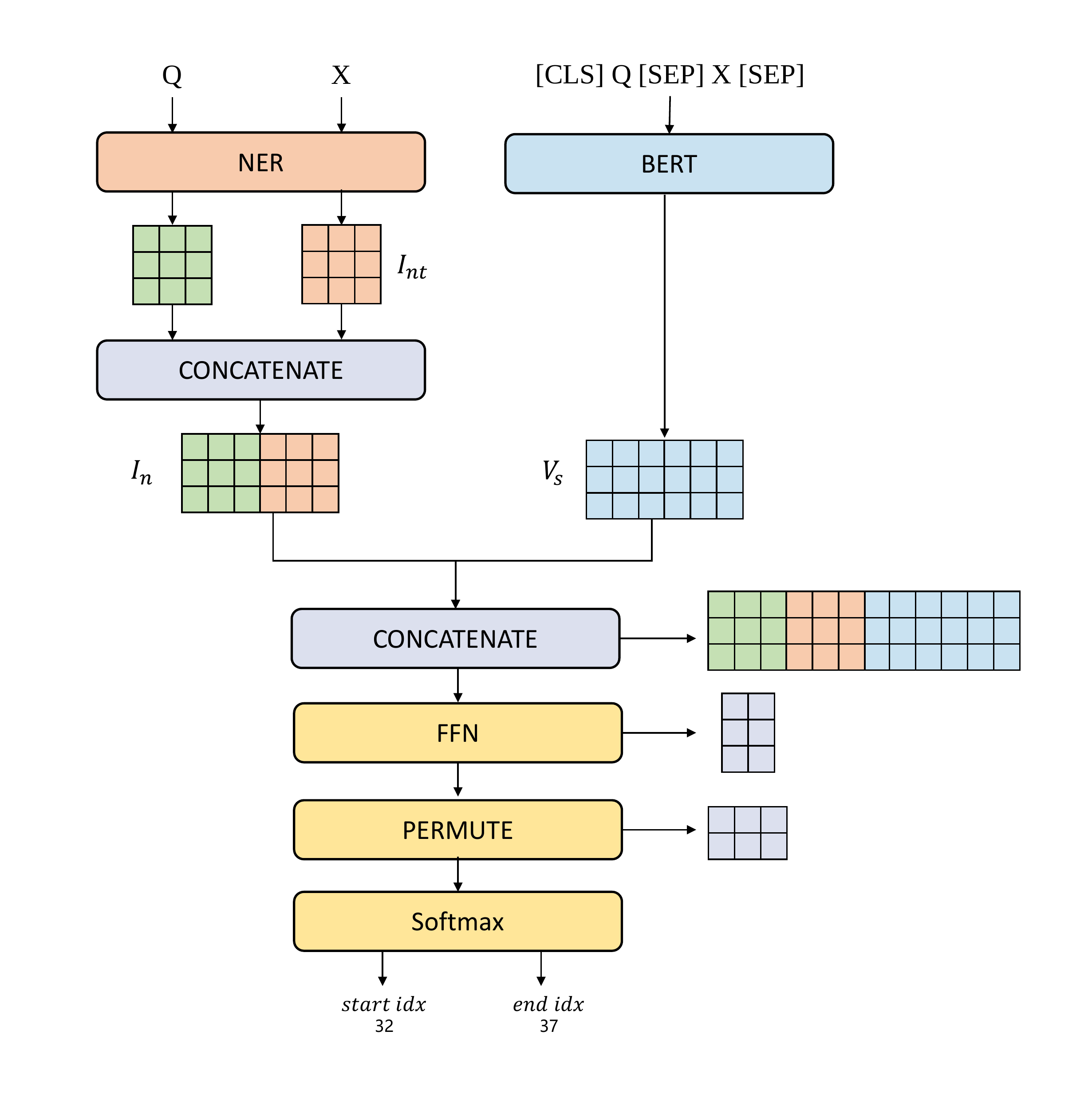}
\caption{The architecture of our proposed model for QA-CTS task}
\label{Fig:Model}
\end{center}
\end{figure}

\section{Question Answering based Clinical Text Structuring}
\label{Sec:Question Answering based Clinical Text Structuring}

Given a sequence of paragraph text $X=<x_1, x_2, ..., x_n>$, clinical text structuring (CTS) can be regarded to extract a key-value pair where key $Q$ is typically a query term such as proximal resection margin and value $V$ is a result of query term $Q$ according to the paragraph text $X$. In some cases, some transformation may be applied to the extracted keys or values.

Generally, researchers solve CTS problem in two steps. Firstly, the answer-related text is pick out. And then several steps such as entity names conversion and negative words recognition are deployed to generate the desired final answer. While final answer varies from task to task, which truly causes non-uniform output formats, finding the answer-related text is a common action among all tasks. Traditional methods regard both the steps as a whole. In this paper, we focus on finding the answer-related substring $Xs = <X_i, X_i+1, X_i+2, ... X_j> (1 <= i < j <= n)$ from paragraph text $X$.

Since BERT~\cite{devlin2018bert} has already demonstrated the usefulness of shared model, we suppose extracting commonality of this problem and unifying the output format will make the model more powerful than dedicated model and meanwhile, for a specific clinical task, use the data for other tasks to supplement the training data.

\section{The Proposed Model for QA-CTS Task}
\label{Sec:The Proposed Model}

In this section, we present an effective model for the question answering based clinical text structuring (QA-CTS). As shown in Fig.~\ref{Fig:Model}, paragraph text $X$ is first passed to a clinical named entity recognition (CNER) model~\cite{jiahui2019chinese,wang2019incorporating} to capture named entity information and obtain one-hot CNER output tagging sequence for query text $I_{nq}$ and paragraph text $I_{nt}$ with BIEOS (Begin, Inside, End, Outside, Single) tag scheme. $I_{nq}$ and $I_{nt}$ are then integrated together into $I_n$. Meanwhile, the paragraph text $X$ and query text $Q$ are organized and passed to contextualized representation model which is pre-trained language model BERT~\cite{devlin2018bert} here to obtain the contextualized representation vector $V_s$ of both text and query. Afterwards, $V_s$ and $I_n$ are integrated together and fed into a feed forward network to calculate the start and end index of answer-related text. Here we define this calculation problem as a classification for each word to be the start or end word.

\subsection{Contextualized Representation of Sentence Text and Query Text}
\label{Sec:Contextualized Representation of Text and Query}

For any clinical free-text paragraph $X$ and query $Q$, contextualized representation is to generate the encoded vector of both of them. Here we use pre-trained language model BERT-base~\cite{devlin2018bert} model to capture contextual information. The text input is constructed as `[CLS] $Q$ [SEP] $X$ [SEP]'. For Chinese sentence, each word in this input will be mapped to a pre-trained embedding $e_i$. To tell the model $Q$ and $X$ is two different sentence, a sentence type input is generated which is a binary label sequence to denote what sentence each character in the input belongs to. Positional encoding and mask matrix is also constructed automatically to bring in absolute position information and eliminate the impact of zero padding respectively. Then a hidden vector $V_s$ which contains both query and text information is generated through BERT-base model.

\subsection{Clinical Named Entity Information}
\label{Sec:Named Entity Information}

Our model integrates clinical named entity information. The clinical named entity recognition (CNER) task aims to identify and classify important clinical terms such as diseases, symptoms, treatments, exams, and body parts from Chinese EHRs. It can be regarded as a sequence labeling task. A CNER model typically outputs a sequence of tags. Each character of the original sentence will be tagged a label following a tag scheme. In this paper we recognize the entities by the model of our previous work~\cite{jiahui2019chinese} but trained on another corpus which has 44 entity types including operations, numbers, unit words, examinations, symptoms, negative words, etc. We denote the sequence for clinical sentence and query term as $I_{nt}$ and $I_{nq}$, respectively.

\subsection{Final Prediction}

The final step is to use integrated representation $H_i$ to predict the start and end index of answer-related text. Here we define this calculation problem as a classification for each word to be the start or end word. We use a feed forward network (FFN) to compress and calculate the score of each word $H_f$ which makes the dimension to $\left\langle l_s, 2\right\rangle$ where $l_s$ denotes the length of sequence.
\begin{equation}
    H_f = FFN(H_i)
\end{equation}

Then we permute the two dimensions for softmax calculation. The calculation process of loss function can be defined as followed.
\begin{equation}
    L = -\sum_{i=1}^{l_s} y_{s_i} log(O_{s_i}) -\sum_{i=1}^{l_s} y_{e_i} log(O_{e_i})
\end{equation}
where $O_s = softmax(permute(H_f)_0)$ denotes the probability score of each word to be the start word and similarly $O_e = softmax(permute(H_f)_1)$ denotes the end. $y_s$ and $y_e$ denotes the true answer of the output for start word and end word respectively.

\subsection{Two-Stage Training Mechanism}
\label{Sec:Two-Stage Training Mechanism}

Two-stage training mechanism is previously applied on bilinear model in fine-grained visual recognition~\cite{lin2015bilinear, gatys2016image, moghimi2016boosted}. Two CNNs are deployed in the model. One is trained at first for coarse-graind features while freezing the parameter of the other. Then unfreeze the other one and train the entire model in a low learning rate for fetching fine-grained features. Inspired by this and due to the large amount of parameters in BERT model, to speed up the training process, we fine tune the BERT model with new prediction layer first to achieve a better contextualized representation performance. Then we deploy the proposed model and load the fine tuned BERT weights, attach named entity information layers and retrain the model.

\section{Experimental Studies}
\label{Sec:Experimental Studies}
In this section, we devote to experimentally evaluating our proposed task and approach. The best results in tables are in bold.

\subsection{Dataset and Evaluation Metrics}
\label{Sec:Dataset and Evaluation Metrics}

Our dataset is annotated based on Chinese pathology reports provided by the Department of Gastrointestinal Surgery, Ruijin Hospital. It contains 17,833 sentences, 826,987 characters and 2,714 question-answer pairs. All question-answer pairs are annotated and reviewed by four clinicians with three types of questions, namely tumor size, proximal resection margin and distal resection margin. These annotated instances have been partitioned into 1,899 training instances (12,412 sentences) and 815 test instances (5,421 sentences). Each instance has one or several sentences. Detailed statistics of different types of entities are listed in Table \ref{Tab:Dataset}.

\begin{table}[!ht]
\begin{center}
\caption{Statistics of different types of question answer instances}
\label{Tab:Dataset}
\begin{tabular}{|c|c|c|}
\hline
Type& Training Set & Test Set \\
\hline
Proximal Resection Margin & 643 & 290\\
Distal Resection Margin & 681 & 270 \\
Tumor Size & 575 & 255 \\
\hline
Total & 1,899 & 815 \\
\hline
\end{tabular}
\end{center}
\end{table}

In the following experiments, two widely-used performance measures (i.e., EM-score~\cite{rajpurkar-etal-2016-squad} and (macro-averaged) F$_1$-score \cite{liu2014strategy,zhou2014correlation}) are used to evaluate the methods.

\subsection{Experimental Settings}
\label{Sec:Experimental Settings}

\begin{table*}[!ht]
\begin{center}
\caption{Comparative Results for Data Integration Analysis (Without mixed-data pre-trained parameters)}
\label{Tab:DIA With}
\begin{tabular}{|c|c|c|c|c|c|c|}
\hline
\multirow{2}{*}{}  & \multicolumn{2}{c|}{Tumor Size} & \multicolumn{2}{c|}{Proximal Resection Margin} & \multicolumn{2}{c|}{Distal Resection Margin} \\
\cline{2-7} & EM-score & F$_1$-score & EM-score & F$_1$-score & EM-score & F$_1$-score \\
\hline
Pure Tumor Size & \textbf{96.27}  & \textbf{96.08} & 0.00 & 17.93  & 0.00 & 21.48  \\
\hline
Pure Proximal Resection Margin & 0.00  & 19.22 & 84.48 & 85.86 & 6.67 & 40.00   \\
\hline
Pure Distal Resection Margin   & 0.00  & 21.18 & 4.65  & 44.83 & 88.33 & 87.41  \\
\hline
Mixed Data  & 95.10 & 94.51 & \textbf{88.45}  & \textbf{88.28} & \textbf{92.41} & \textbf{91.48} \\
\hline
\end{tabular}
\end{center}
\end{table*}

\begin{table*}[!ht]
\begin{center}
\caption{Comparative Results for Data Integration Analysis (Using Mixed-data Pre-trained Parameters)}
\label{Tab:DIA Pretrained}
\begin{tabular}{|c|c|c|c|c|c|c|}
\hline
\multirow{2}{*}{} & \multicolumn{2}{c|}{Tumor Size} & \multicolumn{2}{c|}{Proximal Resection Margin} & \multicolumn{2}{c|}{Distal Resection Margin} \\
\cline{2-7}
& EM-score & F$_1$-score & EM-score & F$_1$-score  & EM-score  & F$_1$-score \\
\hline
Pure Tumor Size & \textbf{96.27} & \textbf{96.08} & 30.86 & 27.93 & 43.52 & 41.48 \\
\hline
Pure Proximal Resection Margin & 71.18 & 61.96 & 85.00  & 87.25 & 69.26 & 70.74 \\
\hline
Pure Distal Resection Margin   & 64.31 & 55.69 & 73.62  & 78.97 & 90.93 & 90.37 \\
\hline
Mixed Data & 95.10 & 94.51 & \textbf{88.45} & \textbf{88.64} & \textbf{92.41} & \textbf{91.48} \\
\hline
\end{tabular}
\end{center}
\end{table*}

To implement deep neural network models, we utilize the Keras library~\cite{chollet2015keras} with TensorFlow~\cite{abadi2016tensorflow} backend. Each model is run on a single NVIDIA GeForce GTX 1080 Ti GPU. The models are trained by the well-known Adam algorithm whose parameters are the same as the default settings except for learning rate set to $5\times10^{-5}$. Batch size is set to 3 or 4 due to the lack of graphical memory. We select BERT-base as the pre-trained language model in this paper. Due to the high cost of pre-training BERT language model, we directly adopt parameters pre-trained by Google in Chinese general corpus. The named entity recognition is applied to both pathology report texts and query texts.

\subsection{Comparison with State-of-the-art Methods}
\label{Sec:Comparison with State-of-the-art Methods}

In this section, we experimentally compare our proposed model with state-of-the-art question answering models (i.e. QANet~\cite{yu2018qanet}) and BERT-Base~\cite{devlin2018bert}. Although BERT has two versions (i.e., BERT-Base and BERT-Large), we only compare our model with the BERT-Base model due to the lack of computational resource. Prediction layer is attached at the end of the original BERT-Base model, and we fine tune it based on our dataset. The named entity integration method is chosen to pure concatenation (Concatenate the named entity information on pathology report text and query text first and then concatenate contextualized representation and concatenated named entity information). Comparative results are summarized in Table~\ref{Tab:Compared With Single Models}.
\begin{table}[!ht]
\begin{center}
\caption{Comparative Results between BERT and Our Proposed Model}
\label{Tab:Compared With Single Models}
\begin{tabular}{|c|c|c|}
\hline
Methods  & EM-score &  F$_1$-score\\
\hline
QANet &  85.45  &  93.62 \\
BERT-Base &  86.20  &  90.06\\
\hline
\textbf{Our Model} &  $\textbf{91.84}$  &  $\textbf{93.75}$\\
\hline
\end{tabular}
\end{center}
\end{table}

Table~\ref{Tab:Compared With Single Models} indicates that our proposed model achieved the best performance both in EM-score and F$_1$-score with EM-score of 91.84\% and F$_1$-score of 93.75\%. QANet outperformed BERT-Base with 3.56\% score in F$_1$-score but underperformed it with 0.75\% score in EM-score. Compared with BERT-Base, our model led to a 5.64\% performance improvement in EM-score and 3.69\% in F$_1$-score. Although our model didn't outperform much with QANet in F$_1$-score (only 0.13\%), our model significantly outperformed it with 6.39\% score in EM-score.

\subsection{Data Integration Analysis}
\label{Sec:Data Integration Analysis}

To investigate how shared task and shared model can benefit, we split our dataset by query types, train our proposed model with different datasets and demonstrate their performance on different datasets.

As indicated in Table~\ref{Tab:DIA With} and Table~\ref{Tab:DIA Pretrained}, using mixed-data pre-trained parameters can significantly improve the model performance than task-specific data trained model. In this case, model trained by mixed data does not have any difference between two table. Except pure tumor size data, the result was improved by 0.52\% score in EM-score, 1.39\% score in F$_1$-score for pure proximal resection margin and 2.6\% score in EM-score, 2.96\% score in F$_1$-score for pure distal resection margin. This proves mixed-data pre-trained parameters can lead to a great benefit for specific task. Meanwhile, the model performance on other tasks which are not trained in the final stage was also improved from around 0 to 60 or 70 percent. This proves that there is commonality between different tasks and our proposed QA-CTS task make this learnable. In conclusion, to achieve the best performance for a specific dataset, pre-training the model in multiple datasets and then fine tuning the model on the specific dataset is the best way.

\section{Conclusion}
\label{Sec:Conclusion and Future Work}

In this paper, we present a question answering based clinical text structuring (QA-CTS) task, which unifies different clinical text structuring tasks and utilize different datasets. A novel model is also proposed to integrate named entity information into a pre-trained language model and adapt it to QA-CTS task. Experimental results on real-world dataset demonstrate that our proposed model competes favorably with strong baseline models in all three specific tasks. The shared task and shared model introduced by QA-CTS task has also been proved to be useful for improving the performance on most of the task-specific datasets.

\section*{Acknowledgment}

We would like to thank Ting Li and Xizhou Hong (Ruijin Hospital) who have helped us very much in data fetching and data cleansing. We would also like to appreciate valuable comments from the anonymous reviewers. This work is supported by the National Key R\&D Program of China for ``Precision Medical Research" (Grant No. 2018YFC0910500), the National Natural Science Foundation of China (Grant No. 61772201), the Special Fund Project for ``Shanghai Informatization Development in Big Data" (Grant No. 201901043).
\end{CJK}

\bibliographystyle{IEEEtran}
\bibliography{IEEEexample}

\begin{thebibliography}{10}
\providecommand{\url}[1]{#1}
\csname url@samestyle\endcsname
\providecommand{\newblock}{\relax}
\providecommand{\bibinfo}[2]{#2}
\providecommand{\BIBentrySTDinterwordspacing}{\spaceskip=0pt\relax}
\providecommand{\BIBentryALTinterwordstretchfactor}{4}
\providecommand{\BIBentryALTinterwordspacing}{\spaceskip=\fontdimen2\font plus
\BIBentryALTinterwordstretchfactor\fontdimen3\font minus
  \fontdimen4\font\relax}
\providecommand{\BIBforeignlanguage}[2]{{%
\expandafter\ifx\csname l@#1\endcsname\relax
\typeout{** WARNING: IEEEtran.bst: No hyphenation pattern has been}%
\typeout{** loaded for the language `#1'. Using the pattern for}%
\typeout{** the default language instead.}%
\else
\language=\csname l@#1\endcsname
\fi
#2}}
\providecommand{\BIBdecl}{\relax}
\BIBdecl

\bibitem{fukuda1998toward}
K.-i. Fukuda, T.~Tsunoda, A.~Tamura, T.~Takagi \emph{et~al.}, ``Toward
  information extraction: identifying protein names from biological papers,''
  in \emph{Pacific Symposium on Biocomputing}, vol. 707, no.~18, 1998, pp.
  707--718.

\bibitem{wang2006linguistic}
Y.~Wang, J.~Patrick, G.~Miller, and J.~O'Halloran, ``Linguistic mapping of
  terminologies to {SNOMED CT},'' in \emph{Semantic Mining Conference on
  {SNOMED CT}}.\hskip 1em plus 0.5em minus 0.4em\relax Citeseer, 2006.

\bibitem{song2015developing}
M.~Song, H.~Yu, and W.-S. Han, ``Developing a hybrid dictionary-based
  bio-entity recognition technique,'' \emph{BMC Medical Informatics and
  Decision Making}, vol.~15, no.~1, p.~S9, 2015.

\bibitem{topaz2016automated}
M.~Topaz, K.~Lai, D.~Dowding, V.~J. Lei, A.~Zisberg, K.~H. Bowles, and L.~Zhou,
  ``Automated identification of wound information in clinical notes of patients
  with heart diseases: Developing and validating a natural language processing
  application,'' \emph{International Journal of Nursing Studies}, vol.~64, pp.
  25--31, 2016.

\bibitem{tan2016development}
H.-J. Tan, R.~Clarke, K.~Chamie, A.~Kaplan, A.~Chin, M.~S.~Litwin, C.~Saigal,
  and A.~S.~Hackbarth, ``Development and validation of an automated method for
  identifying patients undergoing radical cystectomy for bladder cancer using
  natural language processing,'' \emph{Urology Practice}, vol.~4, 11 2016.

\bibitem{senders2019natural}
J.~T. Senders, A.~V. Karhade, D.~J. Cote, A.~Mehrtash, N.~Lamba, A.~DiRisio,
  I.~S. Muskens, W.~B. Gormley, T.~R. Smith, M.~L. Broekman \emph{et~al.},
  ``Natural language processing for automated quantification of brain
  metastases reported in free-text radiology reports,'' \emph{JCO Clinical
  Cancer Informatics}, vol.~3, pp. 1--9, 2019.

\bibitem{bill2014automated}
R.~Bill, S.~Pakhomov, E.~S. Chen, T.~J. Winden, E.~W. Carter, and G.~B. Melton,
  ``Automated extraction of family history information from clinical notes,''
  in \emph{AMIA Annual Symposium Proceedings}, vol. 2014.\hskip 1em plus 0.5em
  minus 0.4em\relax American Medical Informatics Association, 2014, p. 1709.

\bibitem{iqbal2017adept}
E.~Iqbal, R.~Mallah, D.~Rhodes, H.~Wu, A.~Romero, N.~Chang, O.~Dzahini,
  C.~Pandey, M.~Broadbent, R.~Stewart \emph{et~al.}, ``{ADEPt}, a
  semantically-enriched pipeline for extracting adverse drug events from
  free-text electronic health records,'' \emph{PloS one}, vol.~12, no.~11, p.
  e0187121, 2017.

\bibitem{fonferko2019using}
B.~Fonferko-Shadrach, A.~S. Lacey, A.~Roberts, A.~Akbari, S.~Thompson, D.~V.
  Ford, R.~A. Lyons, M.~I. Rees, and W.~O. Pickrell, ``Using natural language
  processing to extract structured epilepsy data from unstructured clinic
  letters: development and validation of the {ExECT} (extraction of epilepsy
  clinical text) system,'' \emph{BMJ Open}, vol.~9, no.~4, p. e023232, 2019.

\bibitem{radford2018improving}
A.~Radford, K.~Narasimhan, T.~Salimans, and I.~Sutskever, ``Improving language
  understanding by generative pre-training,'' \emph{URL https://s3-us-west-2.
  amazonaws. com/openai-assets/researchcovers/languageunsupervised/language
  understanding paper. pdf}, 2018.

\bibitem{peters-etal-2018-deep}
M.~Peters, M.~Neumann, M.~Iyyer, M.~Gardner, C.~Clark, K.~Lee, and
  L.~Zettlemoyer, ``Deep contextualized word representations,'' in
  \emph{Proceedings of the 2018 Conference of the North {A}merican Chapter of
  the Association for Computational Linguistics: Human Language Technologies,
  Volume 1 (Long Papers)}.\hskip 1em plus 0.5em minus 0.4em\relax Association
  for Computational Linguistics, Jun. 2018, pp. 2227--2237.

\bibitem{devlin2018bert}
J.~Devlin, M.-W. Chang, K.~Lee, and K.~Toutanova, ``{BERT}: Pre-training of
  deep bidirectional transformers for language understanding,'' \emph{arXiv
  preprint arXiv:1810.04805}, 2018.

\bibitem{yang2019xlnet}
Z.~Yang, Z.~Dai, Y.~Yang, J.~Carbonell, R.~Salakhutdinov, and Q.~V. Le,
  ``{XLNet}: Generalized autoregressive pretraining for language
  understanding,'' \emph{arXiv preprint arXiv:1906.08237}, 2019.

\bibitem{lee2019biobert}
J.~Lee, W.~Yoon, S.~Kim, D.~Kim, S.~Kim, C.~H. So, and J.~Kang, ``Biobert:
  pre-trained biomedical language representation model for biomedical text
  mining,'' \emph{arXiv preprint arXiv:1901.08746}, 2019.

\bibitem{Kimn2019neural}
D.~{Kim}, J.~{Lee}, C.~H. {So}, H.~{Jeon}, M.~{Jeong}, Y.~{Choi}, W.~{Yoon},
  M.~{Sung}, and J.~{Kang}, ``A neural named entity recognition and multi-type
  normalization tool for biomedical text mining,'' \emph{IEEE Access}, vol.~7,
  pp. 73\,729--73\,740, 2019.

\bibitem{jiahui2019chinese}
J.~{Qiu}, Y.~{Zhou}, Q.~{Wang}, T.~{Ruan}, and J.~{Gao}, ``Chinese clinical
  named entity recognition using residual dilated convolutional neural network
  with conditional random field,'' \emph{IEEE Transactions on NanoBioscience},
  vol.~18, no.~3, pp. 306--315, July 2019.

\bibitem{wang2019incorporating}
Q.~Wang, Y.~Zhou, T.~Ruan, D.~Gao, Y.~Xia, and P.~He, ``Incorporating
  dictionaries into deep neural networks for the chinese clinical named entity
  recognition,'' \emph{Journal of Biomedical Informatics}, vol.~92, p. 103133,
  2019.

\bibitem{lin2015bilinear}
T.-Y. Lin, A.~RoyChowdhury, and S.~Maji, ``Bilinear cnn models for fine-grained
  visual recognition,'' in \emph{Proceedings of the IEEE International
  Conference on Computer Vision}, 2015, pp. 1449--1457.

\bibitem{gatys2016image}
L.~A. Gatys, A.~S. Ecker, and M.~Bethge, ``Image style transfer using
  convolutional neural networks,'' in \emph{Proceedings of the IEEE Conference
  on Computer Vision and Pattern Recognition}, 2016, pp. 2414--2423.

\bibitem{moghimi2016boosted}
M.~Moghimi, S.~J. Belongie, M.~J. Saberian, J.~Yang, N.~Vasconcelos, and L.-J.
  Li, ``Boosted convolutional neural networks.'' in \emph{BMVC}, 2016, pp.
  24--1.

\bibitem{rajpurkar-etal-2016-squad}
\BIBentryALTinterwordspacing
P.~Rajpurkar, J.~Zhang, K.~Lopyrev, and P.~Liang, ``{SQ}u{AD}: 100,000+
  questions for machine comprehension of text,'' in \emph{Proceedings of the
  2016 Conference on Empirical Methods in Natural Language Processing}.\hskip
  1em plus 0.5em minus 0.4em\relax Austin, Texas: Association for Computational
  Linguistics, Nov. 2016, pp. 2383--2392. [Online]. Available:
  \url{https://www.aclweb.org/anthology/D16-1264}
\BIBentrySTDinterwordspacing

\bibitem{liu2014strategy}
Y.~Liu, Y.~Zhou, S.~Wen, and C.~Tang, ``A strategy on selecting performance
  metrics for classifier evaluation,'' \emph{International Journal of Mobile
  Computing and Multimedia Communications (IJMCMC)}, vol.~6, no.~4, pp. 20--35,
  2014.

\bibitem{zhou2014correlation}
Y.~Zhou and Y.~Liu, ``Correlation analysis of performance metrics for
  classifier,'' in \emph{Decision Making and Soft Computing: Proceedings of the
  11th International {FLINS} Conference}.\hskip 1em plus 0.5em minus
  0.4em\relax World Scientific, 2014, pp. 487--492.

\bibitem{chollet2015keras}
F.~Chollet \emph{et~al.}, ``Keras,'' \url{https://keras.io}, 2015.

\bibitem{abadi2016tensorflow}
M.~Abadi, P.~Barham, J.~Chen, Z.~Chen, A.~Davis, J.~Dean, M.~Devin,
  S.~Ghemawat, G.~Irving, M.~Isard \emph{et~al.}, ``Tensorflow: a system for
  large-scale machine learning.'' in \emph{OSDI}, vol.~16, 2016, pp. 265--283.

\bibitem{yu2018qanet}
A.~W. Yu, D.~Dohan, M.-T. Luong, R.~Zhao, K.~Chen, M.~Norouzi, and Q.~V. Le,
  ``Qanet: Combining local convolution with global self-attention for reading
  comprehension,'' \emph{arXiv preprint arXiv:1804.09541}, 2018.

\end{thebibliography}
\end{document}